\ifdefined\pdfminorversion\pdfminorversion=4\fi
\ifdefined\pdfobjcompresslevel\pdfobjcompresslevel=0\fi
\documentclass[letterpaper,10pt,conference]{ieeeconf}

\IEEEoverridecommandlockouts
\usepackage{booktabs}
\usepackage{graphicx}
\usepackage{microtype}
\usepackage{amsmath,amssymb}
\usepackage{multirow}
\usepackage{cite}
\usepackage[table]{xcolor}
\usepackage{xurl}
\usepackage{flushend}
\definecolor{oursblue}{RGB}{232,242,250}

\newcommand{\papertitle}{FutureRay: Control-Aligned Future Range for Agile Quadruped Navigation}

\title{\LARGE \bf \papertitle}
\author{Tianhao Zang$^{1,2}$, Shanze Wang$^{4}$, Ziqian Wang$^{3}$, Liyou Luo$^{5}$,\\
Zihan Liu$^{3}$, Xingjian Xie$^{3}$, Wei Zhang$^{1,*}$
\thanks{$^{1}$College of Information Science and Technology, Eastern Institute of Technology, Ningbo, P. R. China.}
\thanks{$^{2}$University of Maryland, USA.}
\thanks{$^{3}$School of Computer Science, University of Nottingham, UK.}
\thanks{$^{4}$The Hong Kong Polytechnic University, Hong Kong.}
\thanks{$^{5}$Xi'an Jiaotong-Liverpool University, China.}
\thanks{$^{*}$Corresponding author. Email: \texttt{zhw@eitech.edu.cn}.}}

\begin{document}
\maketitle
\thispagestyle{empty}
\pagestyle{empty}

\begin{abstract}
Moving obstacles can block a previously clear route while a quadruped robot executes a motion command. We investigate whether predicting changing clearance improves navigation when motion selection accounts for the robot footprint and the time needed to react and brake. We present FutureRay, which predicts ranges across viewing directions and future times, together with encounter risk, from depth-derived range history and observable robot motion. Training emphasizes near-term clearance and penalizes errors that overstate available space. A local planner queries the same forecast for candidate headings and combines it with current observations to check clearance around the robot footprint. Model-based reaction--braking limits guide speed selection, and the resulting velocity commands are passed to a fixed locomotion policy. In paired evaluations on 60 static and dynamic simulation scenes, FutureRay achieves 93.3\% completion, compared with 75.0\% for current-range persistence and 80.0\% for Cartesian Kalman rollout, with perception, planning, and locomotion held fixed. FutureRay also records fewer collisions than both baselines. Qualitative trials on a physical quadruped show avoidance initiated while an obstacle is approaching the route, followed by renewed goal progress. These results show that joint range and encounter-risk prediction can improve obstacle avoidance without retraining the locomotion policy.
\end{abstract}
\section{Introduction}

\begin{figure}[t]
  \centering
  \includegraphics[width=\columnwidth]{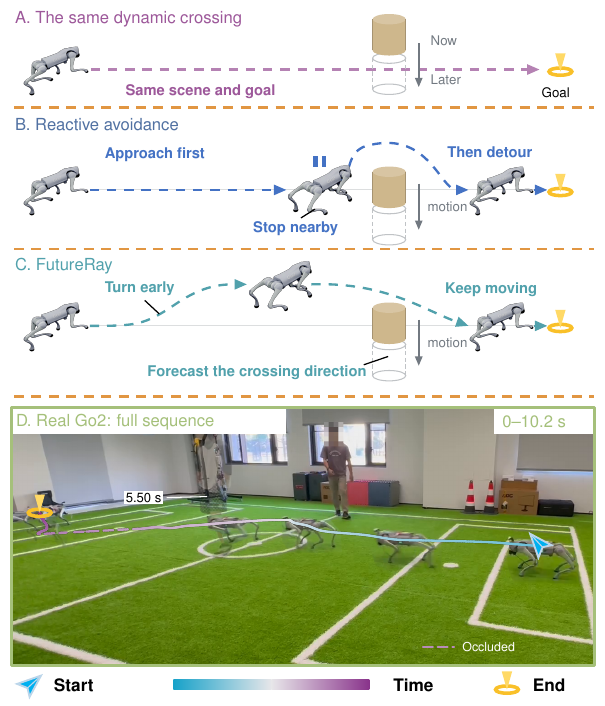}
  \caption{Future sensor space for anticipatory avoidance. The schematics
  contrast a response to current clearance with a detour initiated from a
  forecast of an approaching crossing. The physical quadruped robot sequence illustrates
  onboard avoidance and subsequent progress toward the goal.}
  \label{fig:preview-control}
\end{figure}

Learned locomotion enables quadrupeds to traverse uneven terrain~\cite{Rudin2022,Miki2022}, while perceptive navigation supports movement through complex obstacle courses~\cite{Hoeller2024}. Moving obstacles introduce an additional challenge: a currently clear passage can become occupied as the robot moves. The robot needs space to accommodate its body and time to react and brake, so waiting until an obstacle is close can leave insufficient room to avoid it. Heading and speed selection should therefore consider how available clearance will change, together with the robot footprint and stopping distance. The crossing scenario in Fig.~\ref{fig:preview-control} illustrates this need for anticipation.

Existing work estimates directional ranges from depth images for collision avoidance~\cite{AgileButSafe}. Other methods incorporate LiDAR history into locomotion or navigation~\cite{Wang2025Omni,Yuan2025REASAN}; auxiliary prediction of the next hidden state can also shape recurrent navigation representations during training~\cite{Zhu2026Latent}. Prediction also enters decisions directly: VOP-Nav predicts safe-velocity regions from LiDAR history for a navigation policy~\cite{Wu2026VOP}, while Ginerica et al. forecast depth-derived occupancy grids and extract obstacle information for local planning~\cite{Ginerica2024Vision}. We instead predict ranges across viewing directions and future times. A local planner queries these forecasts to check space for the robot footprint along candidate headings and compute directional speed limits. A separate risk score predicts whether an obstacle will be nearby over the prediction horizon and provides an additional input for candidate action evaluation. We ask whether these learned range and risk predictions improve obstacle avoidance and goal reaching compared with repeating current ranges or extrapolating observed geometry with a Cartesian Kalman filter, with perception, planning, and locomotion held fixed.

For this planner, the direction of a predicted range indicates where clearance may change, while its forecast time distinguishes imminent changes from later ones. Metric ranges support footprint checks and speed selection based on modeled reaction and braking distances, following classical reasoning about collision avoidance and stopping feasibility~\cite{Fox1997,Fiorini1998VO}. Prediction errors can alter these decisions: placing a nearby obstacle too far away overstates the available space and can favor overly aggressive motion. REASAN already penalizes overestimation more heavily than underestimation when learning current range estimates~\cite{Yuan2025REASAN}. For future ranges, near-term errors can affect the next motion decision. We therefore give greater training weight to near-term range errors and errors involving nearby obstacles, with larger penalties for overestimation. This follows prior work that evaluates prediction errors by their effects \mbox{on downstream decisions~\cite{Huang2022TIP,McAllister2022CAPO}}.

We present FutureRay, which uses depth-derived range history and observable robot motion to predict future range observations in the forward field of view, together with short-horizon encounter risk. A local planner queries the same forecast for candidate headings and combines it with current observations to check space around the robot footprint. It separately queries radial ranges around each heading and uses a reaction--braking model to construct candidate velocities. These candidates are evaluated using goal progress, directional clearance, and encounter risk. Magnitude and rate limits are applied before the selected commands reach the fixed quadruped locomotion policy (Fig.~\ref{fig:method-pipeline}).

To answer the research question, we compare FutureRay with current-range persistence and Cartesian Kalman rollout in paired simulation scenes, holding perception, planning, and locomotion fixed. Both baselines derive risk from their range fields, so the comparison evaluates the combined range and risk predictions. Across static and dynamic scenes, FutureRay improves completion and mean minimum clearance while reducing collisions relative to both baselines (Table~\ref{tab:high-speed-main}). Compared with Kalman, FutureRay completes more routes in ordinary dynamic scenes and the same number in high-dynamic scenes (Fig.~\ref{fig:extended-evidence}(b)). FutureRay also achieves the highest overall completion in a comparison with six author-released navigation methods under a common quadruped transfer protocol (Table~\ref{tab:recent-transfer}). Qualitative hardware trials further show a physical quadruped initiating avoidance while an obstacle approaches the route and then resuming progress toward the goal (Fig.~\ref{fig:preview-control}). Our contributions are:

\begin{itemize}
  \item We introduce joint prediction of future directional ranges and encounter risk for local planning, with larger training penalties for range overestimation.
  \item We link shared forecasts to a fixed quadruped locomotion policy through footprint corridor checks and speed selection under a reaction--braking model.
  \item We demonstrate improved navigation in 60 paired simulation scenes, reaching 93.3\% completion (Table~\ref{tab:high-speed-main}), and complement these results with comparisons of released navigation models and qualitative hardware trials.
\end{itemize}
\section{Related Work}

\subsection{Perceptive and Temporal Navigation}

Perceptive locomotion connects onboard geometry to terrain negotiation and
collision avoidance~\cite{Miki2022,Agarwal2023,Hoeller2024,AgileButSafe}.
ABS provides a depth-to-ray interface for agile and recovery policies,
while Omni-Perception and REASAN incorporate temporal LiDAR information
within locomotion or modular navigation systems~\cite{Wang2025Omni,Yuan2025REASAN}.
Temporal encoders, predictive BEV features, and recurrent world models
represent environmental change for navigation~\cite{Lai2026DARE,Jiang2026BEVOSP,Shanks2025DreamerNav}.
Predictive training with latent imagination supervises a recurrent navigation
state through an auxiliary branch used only during
training~\cite{Zhu2026Latent}. These approaches establish the value of
observation history and predictive features; their representation determines
how anticipation enters control. FutureRay places this connection in an
explicit metric range forecast available at inference, where a body-velocity
planner can query it with geometric constraints.

\subsection{Predictive Geometry and Collision Avoidance}

Future occupancy and LiDAR prediction represent evolving geometry without
requiring object identities~\cite{Mann2022,Deng2020Temporal}; sensor-space
safety envelopes likewise describe nearby free-space
boundaries~\cite{Ancha2021SafetyEnvelope}. Closest to our geometric interface,
Ginerica et al. predict depth-derived occupancy grids, cluster the predicted
geometry, and supply obstacle points to a modified Dynamic Window
planner~\cite{Ginerica2024Vision}. FutureRay retains angular metric ranges
as the planner's query representation, avoiding occupancy reconstruction
and clustering in its online geometric interface.

The Dynamic Window Approach and Velocity Obstacles connect geometric
evidence to stopping feasibility and future collision-inducing
velocities~\cite{Fox1997,Fiorini1998VO}. Learned motion forecasts can also
enter model-predictive-control constraints~\cite{Zhang2025Future}.
VOP-Nav learns a Velocity-Obstacle-derived safe-velocity region directly
from LiDAR history~\cite{Wu2026VOP}, exposing the prediction in velocity
space. FutureRay, a control-aligned future-range interface for quadruped
navigation, preserves bearing and forecast time in a shared nominal range
field. This representation lets the planner query footprint corridors and
directional speed limits while using forecast order to evaluate proximity
and clearance trends.

\subsection{Control-Aware Prediction}

Task-informed prediction and control-aware objectives evaluate prediction
errors through their consequences for downstream
decisions~\cite{Huang2022TIP,McAllister2022CAPO}. Asymmetric prediction
losses also appear in depth-derived occupancy forecasting for
navigation~\cite{Ginerica2024Vision}. These works motivate tailoring the
objective to the information consumed by control. Perceptive forward-dynamics
models address the complementary quantity of robot response and failure
under candidate actions~\cite{Roth2025}. FutureRay applies control-aware
supervision to angular range through near-term and nearby-surface weighting,
clearance-overestimation penalties, regional minima, and proximity-trigger
agreement. Its shared-planner evaluation tests the resulting learned
range--risk packet as the interface between future sensor observations and
physical velocity selection.
\section{Preliminaries}

\subsection{Problem Formulation}

At time $t$, we map forward depth $D_t$, observable robot state $e_t$, and planar goal $g$ to the body-frame velocity command $u_t=[v_x,v_y,\omega]$. We select motion toward the goal while accounting for the robot footprint and moving obstacles. The locomotion policy and its joint-action interface remain fixed.

We represent observed geometry by a horizontal range field $R_t\in(0,R_{\max}]^{J}$ at ordered bearings $\{\phi_j\}_{j=1}^{J}$. Current-range persistence repeats this field over the planning horizon. FutureRay instead predicts future ranges and encounter risk:
\begin{equation}
 P_t=\bigl(\widehat R_{t+1:t+K},q_t\bigr),\qquad
 \widehat R\in(0,6]^{K\times J},\quad q_t\in[0,1],
 \label{eq:packet}
\end{equation}
where $R_{\max}=6$~m, $K=10$, $J=90$, and successive forecasts are 0.05~s apart. The scalar $q_t$ estimates whether an obstacle will be nearby within the horizon; the training label is specified below. After validity checks, the planner uses these predictions with current depth-derived ranges, the goal, and observable robot state. Simulator obstacle poses, identities, velocities, contacts, and future states are excluded from online inputs. Geometry and contact data are used only for offline supervision or evaluation.

\section{FutureRay Approach}

\subsection{Control-Aligned Future Sensor Space}

\begin{figure*}[t]
  \centering
  \includegraphics[width=\textwidth]{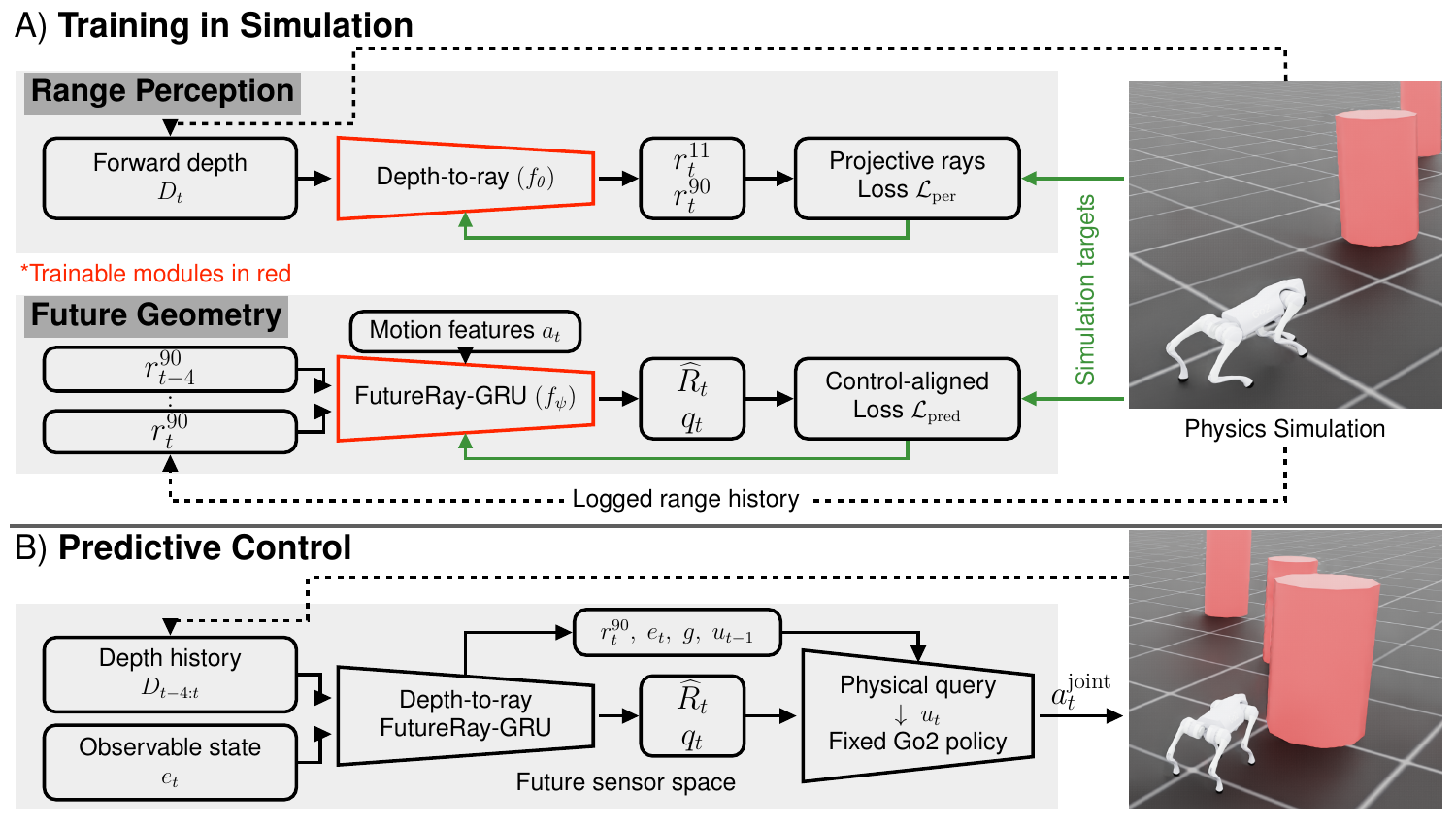}
  \caption{FutureRay training and control. Depth-derived range history and
  observable ego motion produce a future-range field and encounter risk.
  Current geometry, physical constraints, and temporal utility determine
  candidate body velocities. Command-history regularization and rate limits
  precede the fixed quadruped locomotion policy. Supervision is confined to the
  offline training path. Here, $r_t^{90}=R_t$, and $\widehat R_t$ denotes
  the future-range field issued at time $t$.}
  \label{fig:method-pipeline}
\end{figure*}

Fig.~\ref{fig:method-pipeline} summarizes FutureRay. Learned perception converts forward depth into directional ranges. A recurrent predictor combines recent range observations with robot motion to estimate future ranges and encounter risk. The planner uses these predictions, current ranges, and the goal to evaluate candidate motions, applies physical command limits, and sends the body-frame velocity to the fixed locomotion policy.

The camera provides $160\times90$ optical-axis depth images with a $102^{\circ}$ field of view, clipped to 0.1--6.0~m. A ResNet-18~\cite{He2016ResNet} processes the clipped $\log_2 D_t$ images and predicts two log-range outputs: 11 rays and 90 rays over $[-45^{\circ},45^{\circ}]$. Both outputs are supervised by offline geometry:
\begin{equation}
 \mathcal L_{\rm per}=\operatorname{MSE}(\hat y^{11},\log_2 y^{11})+
 \operatorname{MSE}(\hat y^{90},\log_2 y^{90}).
 \label{eq:perception-loss}
\end{equation}
Here, $y^{11}$ and $y^{90}$ are geometric range targets, and $\hat y^{11}$ and $\hat y^{90}$ are predicted log ranges. The 11-ray head preserves the perception interface of ABS~\cite{AgileButSafe}, while forecasting and planning use the 90-ray head.

The predictor takes five range fields sampled at 20~Hz and six observable context features:
\begin{equation}
 a_t=[\bar v_x,\bar v_y,\sin\psi,\cos\psi,
       \tau_{\min}/10.5,\dot r_{\max}],
 \label{eq:aux}
\end{equation}
Here, $\bar v_x,\bar v_y$ are body velocities divided by 1~m/s and clipped to $[-3,3]$, and $\psi$ is robot yaw. Consecutive range fields provide per-bearing closing rates clipped to 0--6~m/s and range-to-closing-rate ratios clipped at 10.5~s. Their maximum and minimum give $\dot r_{\max}$ and the ray-based time-to-contact estimate $\tau_{\min}$, respectively; a nonclosing ray is assigned 10.5~s. A $90\!\rightarrow\!256\!\rightarrow\!128$ encoder with layer normalization and GELU activation processes each range field. A GRU~\cite{Cho2014GRU} with 128 hidden units encodes the five-frame sequence. A context MLP combines its output with $a_t$, and two heads produce 900 future ranges and a sigmoid encounter score. The predictor contains 504,709 trainable parameters.

Training targets are planar radial ranges measured from the robot root at later logged positions, with bearings relative to the yaw at each position. FutureRay thus predicts future sensor observations from range history and observable robot motion. The same nominal forecast is used to evaluate all candidate headings; it is not recomputed for the future position and yaw associated with each candidate.

We weight prediction errors according to their relevance to avoidance, emphasizing near-term forecasts, center bearings, and nearby obstacles. For smooth-$L_1$ penalty $h_{0.1}$, the ray weights are
\begin{equation}
 \begin{aligned}
 w_{kj}\propto{}&3^{[k\leq3]}2^{[j\in\mathcal C]}
 (1+5[R_{kj}<0.85])\\
 &\times(1+2[R_{kj}<1.2,\widehat R_{kj}>R_{kj}]),
 \end{aligned}
 \label{eq:ray-weight}
\end{equation}
where $R_{kj}$ is the target at forecast step $k$ and bearing $j$, $\mathcal C=\{41,\ldots,50\}$ contains the center bearings, and brackets denote an indicator. We normalize the weights to unit mean over all forecast entries in the minibatch. The ray loss $\mathcal L_{\rm ray}$ averages $w_{kj}h_{0.1}(\widehat R_{kj}-R_{kj})$ over these entries and is combined with clearance, proximity-trigger, and risk losses:
\begin{equation}
 \mathcal L_{\rm pred}=\mathcal L_{\rm ray}
 +0.5\mathcal L_{\rm clr}+0.5\mathcal L_{\rm trig}
 +0.25\mathcal L_{\rm risk}.
 \label{eq:predictor-loss}
\end{equation}
The clearance loss compares predicted and target soft minimum ranges over the horizon in the full, left, center, and right regions, with a larger penalty for overestimation. The trigger loss uses binary cross-entropy between a smooth proximity test on predicted regional minima and an analytic target label. This label is positive when the target center minimum is below 0.8~m or the overall minimum is at most 0.538~m, both taken over the horizon. The separate risk head is trained by binary cross-entropy to predict whether any future geometry range falls below 1.65~m.

\subsection{Querying Future Sensor Space for Control}

The planner evaluates eleven translation headings from $-45^{\circ}$ to $45^{\circ}$ in the current body frame; yaw rate follows a separate goal-alignment rule. It uses the learned forecast and risk when the minimum observed range is below 5~m. Otherwise, current-range persistence supplies the planning field. Each forecast slice is queried in a common body-centered polar frame, using the shared forecast described above. For candidate heading $\theta_m$, each predicted range has longitudinal and lateral coordinates
\begin{equation}
 \ell_{mkj}=\widehat R_{kj}\cos(\phi_j-\theta_m),\qquad
 b_{mkj}=\widehat R_{kj}\sin(\phi_j-\theta_m).
 \label{eq:corridor}
\end{equation}
The straight corridor along that heading contains samples satisfying $\ell_{mkj}>0$ and $|b_{mkj}|\leq r_{\rm robot}+m_c$, where $r_{\rm robot}=0.35$~m and $m_c=0.12$~m account for the robot radius and clearance margin. The minimum longitudinal distance among these samples defines the corridor clearance for each heading and forecast step; an empty corridor returns $R_{\max}$. At the first step, the planner takes the minimum of this value and the corresponding current-geometry clearance. A clearance of at most 0.58~m in at least two of the first three steps indicates predicted proximity. Current footprint proximity is checked separately.

Speed selection uses radial distance rather than the longitudinal corridor clearance. Let $c_{mk}$ be the minimum range in a narrow angular window around heading $m$ at forecast step $k$. At the first step, the planner takes its minimum with the corresponding current range. With usable distance $d_{mk}=\max(c_{mk}-r_{\rm robot}-m_c,0)$, reaction time $t_r=0.18$~s, and braking acceleration $a_b=2.0$~m/s$^2$, the speed bound is
\begin{equation}
 v_{{\rm safe},mk}=-a_bt_r+\sqrt{(a_bt_r)^2+2a_bd_{mk}}.
 \label{eq:safe-speed}
\end{equation}
This model assumes a reaction interval followed by constant braking. The lower quartile of the bounds over the horizon sets the initial speed along each candidate heading, before lateral yielding and component and rate limits are applied. Predicted proximity can reduce forward motion and favor lateral yielding. Except during goal realignment, a current-proximity violation in the selected corridor requests braking. Nominal forward and lateral limits are 1.9 and 0.8~m/s; risk-dependent speed profiles and command rate limits further shape the output.

Candidate utility combines goal progress, lower-quartile directional clearance, the learned risk score weighted toward forward headings, and current obstacle-side pressure. It also includes a time-discounted proximity term, $O_m=\sum_k\alpha_k\exp(-c_{mk}/0.75)$, where $\alpha_k\propto\exp(-k\Delta t/0.35)$ and $\sum_{k=1}^{K}\alpha_k=1$. Here, $\Delta t=0.05$~s; 0.75~m and 0.35~s are the range and time scales. For each heading $m$, the clearance trend is the mean of $c_{mk}$ over the last three forecast steps minus its mean over the first three, divided by $R_{\max}$. Together, the early-proximity check, temporal discounting, and clearance trend distinguish imminent obstruction from later changes.

Penalties on heading and command changes use the previous command to discourage abrupt switches between replans. Goal realignment steers the robot back toward a target outside the $[-45^{\circ},45^{\circ}]$ range field. Forward acceleration and deceleration, lateral acceleration, and yaw acceleration are limited before the command enters the fixed locomotion-policy observation. As new observations enter the 20-Hz history, the planner updates command selection. The stateful locomotion policy produces 12 joint actions, ordered by leg, at 50~Hz.
\section{Simulation Setup and Training}

All simulation studies use the quadruped asset and a fixed stateful locomotion policy with 45-dimensional observations. Physics runs at 200~Hz, joint control at 50~Hz, and depth-history sampling at 20~Hz. The three primary prediction methods share locomotion weights, depth-to-range perception, and the candidate planner. Experiments use Isaac Sim 6.0.1, Isaac Lab 6.1.17, and an NVIDIA RTX A5000. An auxiliary Cartesian filter checks the validity of FutureRay predictions but does not provide ranges for candidate motions.

The training set contains 81,400 samples from 814 quadruped navigation scenes: 50.9\% static, 29.5\% ordinary dynamic, and 19.7\% high-dynamic. A disjoint validation set contains 10,200 samples from 102 scenes with the same approximate 51:29:20 composition. Each sample comprises five 90-ray observations, six context features, ten future range fields evaluated at later logged poses, and an encounter label.

The curriculum starts with static layouts, then uses the fixed scene mixture. Static scenes include sparse obstacle layouts and staggered slalom corridors. Dynamic crossings are timed to coincide with nominal robot arrival at 1.8~m/s. High-dynamic scenes increase crossing speed and obstacle density; obstacle motion is not designed to pursue the robot or force collisions. These scenes cover both stationary clearance and geometry that changes within the prediction horizon.

Both learned modules are trained with Adam, using a learning rate of $10^{-3}$ and batch size 128 for perception, and $2\times10^{-4}$ and 256 for FutureRay. Predictor training runs for up to 40 epochs with early-stopping patience of eight epochs; validation selects epoch 28. Validation range MAEs are 0.225~m over all forecast entries and 0.157~m over all bearings in the first three steps. The MAE between predicted and target minima over the center bearings and full horizon is 0.192~m. Sensitivity and specificity of the analytic proximity trigger defined above are 92.9\% and 98.8\%, respectively. The risk activation threshold is selected on the predictor validation set to maximize sensitivity while limiting false positives in static scenes to 1\%.

\begin{figure*}[t]
  \centering
  \includegraphics[width=\textwidth]{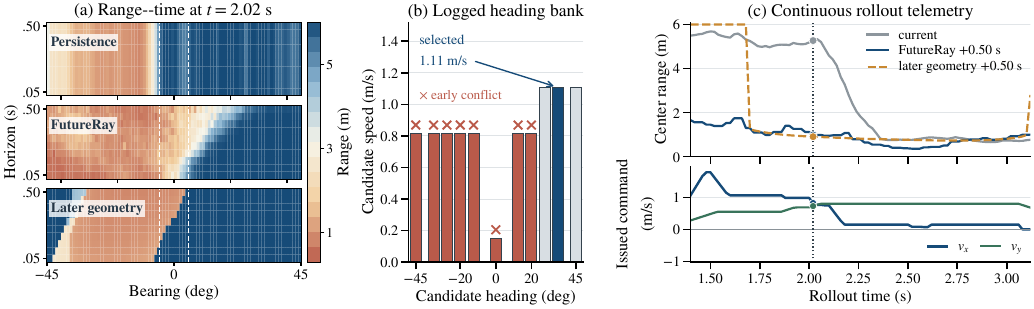}
  \caption{Forecast-to-control diagnostic from one high-dynamic quadruped rollout.
  (a) Current-range persistence, the learned nominal forecast, and offline
  later-pose geometry at $t=2.02$~s share bearing--time axes. The center range
  is 5.27~m; its $+0.50$-s prediction and later geometry are 0.99 and 0.91~m.
  (b) The logged heading bank marks early corridor proximity with crosses
  and selects $+35^\circ$ at 1.11~m/s. (c) Synchronized center-range and
  issued body-command traces show the encounter; issued commands include
  component and rate limits. Later geometry is evaluation-only and uses
  the executed future robot-root positions and yaw angles.}
  \label{fig:control-diagnostic}
\end{figure*}

The subsequent scene-paired evaluation compares current-range persistence, Cartesian Kalman prediction~\cite{Kalman1960}, and FutureRay through the same physical planner.
\section{Performance Validation in Simulation}

\subsection{Protocol and Baselines}

The navigation cohort contains 60 scenes previously used for controller
comparisons, with 20 each in the static, ordinary dynamic, and high-dynamic
strata. These instances differ from predictor-training and
predictor-validation scenes but share their procedural layout and motion
families. The nine methods compared here use fixed weights and parameters with one
common simulator seed for paired evaluation. Methods share start, goal,
obstacle geometry and open-loop motion, raw camera input, and locomotion.
Completion requires arrival within 0.65~m; any positive obstacle-owned
contact force during a physics substep counts as collision. Base contact
marks a fall, and the time limit is 12~s. Outcome flags may overlap.
Forward commands extend to
1.9~m/s. Clearance accounts for the 0.35-m footprint. Body-forward speed
is averaged after 2~s, using the whole episode if that window is empty;
mean path speed in Fig.~\ref{fig:extended-evidence}(a) equally averages
episode traveled distance divided by duration, including failures.

Reactive persistence repeats the current learned ranges over the horizon,
retaining the ABS-compatible perception interface~\cite{AgileButSafe}.
The analytic comparator forecasts depth surfaces with a causal Cartesian
Kalman rollout~\cite{Kalman1960}. Both derive risk analytically from their
range fields; FutureRay predicts range and encounter risk jointly.
All three interfaces use the same physical candidate planner and frozen
quadruped locomotion policy. The experimental variable is the range--risk packet
supplied to this common velocity-command layer.

Six author-released navigation methods retain their modules, preprocessing,
and speed targets when transferred to the quadruped robot. Causal adapters map forward depth
to their inputs and their outputs to planar commands; a fixed visible-body
mask removes self-points. These comparisons measure released-checkpoint
transfer under the same command limits and evaluation criteria.

Binary outcomes are reported through rates and effect sizes.
For fixed configurations and simulator seed, 50,000
within-stratum paired bootstrap draws yield marginal 95\% intervals over
scenes. All episodes, including failures, receive equal weight.

\subsection{Navigation Performance}

FutureRay leads the shared-planner comparison in completion and minimum
clearance (Table~\ref{tab:high-speed-main}). Because perception, candidate
planning, and locomotion are held fixed, these gains isolate the effect of
the range--risk packet supplied to the planner. Mean minimum clearance
reaches 0.534~m, increasing by 0.201~m over persistence
(95\% CI [0.155, 0.248]) and by 0.079~m over Kalman
(95\% CI [0.040, 0.120]).

The interface improves completion while reducing collisions. Completion
increases by 18.3 percentage points over persistence and 13.3 points over
Kalman. Collision counts fall from 12 to 1 and from 7 to 1, reductions of
91.7\% and 85.7\%, respectively. Mean body-forward speed is 1.164 versus
1.321~m/s for persistence. Relative to Kalman, the speed difference is
0.043~m/s (95\% CI [$-0.029$, 0.121]).

\begin{table}[t]
\vspace*{6pt}\caption{Shared-planner quadruped navigation comparison ($n=60$).
S/C: completion/collision (\%); $d$: mean minimum clearance (m);
$v_x$: body-forward speed after 2~s (whole episode if shorter).
All tables shade FutureRay and bold metric bests, including ties.}
\label{tab:high-speed-main}
\centering
\footnotesize
\setlength{\tabcolsep}{3pt}
\begin{tabular}{@{}lrrrr@{}}
\toprule
Method & S $\uparrow$ & C $\downarrow$ & $d\uparrow$ & $v_x\uparrow$ (m/s) \\
\midrule
Reactive persistence & 75.0 & 20.0 & 0.333 & \textbf{1.321} \\
Kalman rollout & 80.0 & 11.7 & 0.455 & 1.121 \\
\rowcolor{oursblue} FutureRay (ours) & \textbf{93.3} & \textbf{1.7} & \textbf{0.534} & 1.164 \\
\bottomrule
\end{tabular}
\end{table}

FutureRay ranks first in completion and mean path speed in the comparison
with six transferred navigation methods (Fig.~\ref{fig:extended-evidence}(a)).
REASAN has the next-highest completion at 85.0\%
(Table~\ref{tab:recent-transfer}). Across all nine methods, FutureRay combines
the highest completion rate (93.3\%) with the lowest collision rate (1.7\%).
CrowdSurfer has a 3.3\% collision rate but completes no routes, with
59 timeouts.

\begin{table}[t]
\caption{Released-checkpoint transfer to the quadruped robot on the same 60 scenes.
S/C/T: completion/collision/timeout (\%);
$v_x$: body-forward speed (m/s), defined as in Table~\ref{tab:high-speed-main};
$\mathrm{S}_{\rm HD}$: completion (\%) on 20 high-dynamic scenes.}
\label{tab:recent-transfer}
\centering
\footnotesize
\setlength{\tabcolsep}{2.5pt}
\begin{tabular}{@{}lrrrrr@{}}
\toprule
Method & S $\uparrow$ & C $\downarrow$ & T $\downarrow$ & $v_x\uparrow$ & $\mathrm{S}_{\rm HD}\uparrow$ \\
\midrule
SRU~\cite{Yang2025SRU} & 56.7 & 30.0 & 13.3 & 1.207 & 20.0 \\
CE-Nav~\cite{Yang2026CENav} & 66.7 & 33.3 & 3.3 & \textbf{1.235} & 80.0 \\
REASAN~\cite{Yuan2025REASAN} & 85.0 & 13.3 & 1.7 & 1.231 & 80.0 \\
NavRL (planar)~\cite{Xu2025NavRL} & 70.0 & 5.0 & 25.0 & 0.793 & 70.0 \\
CrowdSurfer~\cite{Kumar2025CrowdSurfer} & 0.0 & 3.3 & 98.3 & 0.632 & 0.0 \\
SanD-Planner~\cite{Wang2026SanD} & 73.3 & 11.7 & 15.0 & 0.847 & 65.0 \\
\rowcolor{oursblue} FutureRay (ours) & \textbf{93.3} & \textbf{1.7} & \textbf{0.0} & 1.164 & \textbf{100.0} \\
\bottomrule
\end{tabular}
\end{table}

\subsection{Performance Across Scene Types}

\begin{figure*}[t]
  \centering
  \includegraphics[width=\textwidth]{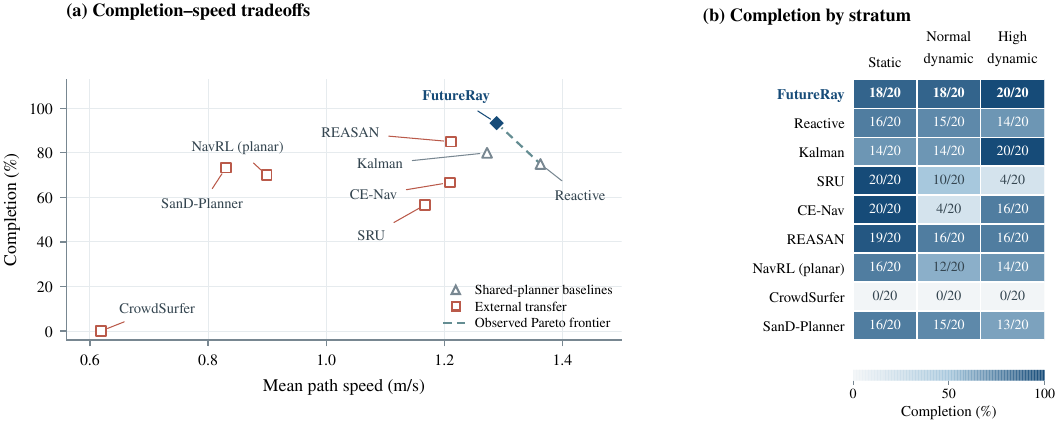}
  \caption{Unified navigation comparison.
  (a) FutureRay, two shared-planner baselines, and six transferred methods
  on the same 60 paired scenes. The diamond denotes FutureRay; triangles
  and squares denote shared-planner baselines and transferred methods.
  Mean path speed equally averages episode distance/duration, including
  failures, and differs from the body-forward speed in the tables.
  Dashed segments mark the observed Pareto frontier, maximizing both axes.
  (b) Completion counts for the primary interface and released-method
  comparisons, with 20 scenes in each stratum. Color encodes completion
  percentage on a common scale.}
  \label{fig:extended-evidence}
\end{figure*}

FutureRay completes all 20 high-dynamic routes, matching Kalman
(Fig.~\ref{fig:extended-evidence}(b)). CE-Nav and REASAN lead the transferred
methods with 16 completions each. In ordinary dynamic scenes, FutureRay
completes 18 routes, the highest count among all methods, versus 14 for
Kalman. It also completes 18 static routes; SRU and CE-Nav each complete 20.
The completion advantage over persistence spans static bypass and dynamic
crossings.

\subsection{Prediction-to-Control Behavior}

In Fig.~\ref{fig:control-diagnostic}, the current center range is large,
while the forecast predicts a decrease within half a second. The corridor
query flags early proximity for the forward heading and selects a lateral
candidate with forward motion. Synchronized traces
show the prediction error and command-rate limiting.

\begin{figure*}[t]
  \centering
  \includegraphics[width=\textwidth]{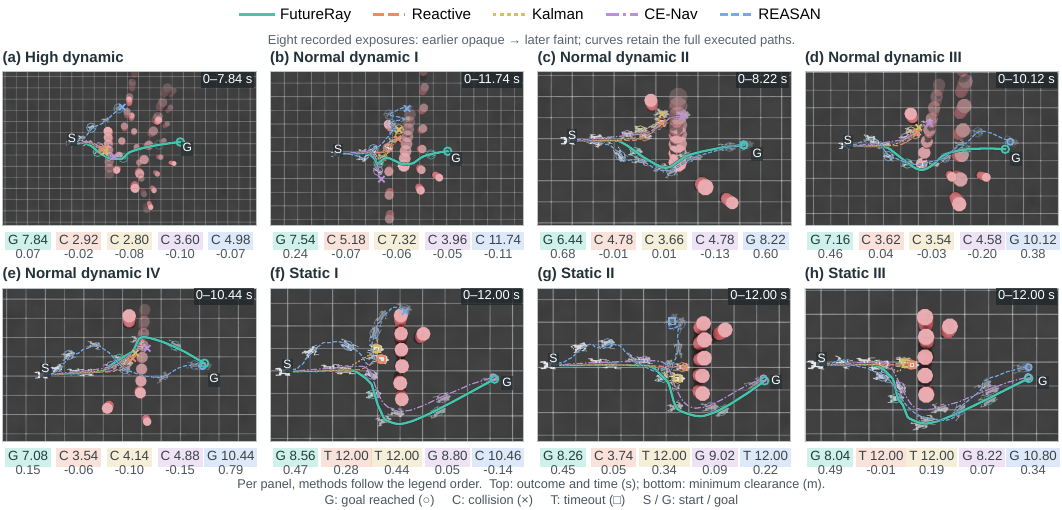}
  \caption{Eight selected Isaac Sim comparisons: (a)--(e) dynamic crossings and (f)--(h) static bypass. Each panel overlays eight synchronized exposures, fading from early to late, and complete executed trajectories of FutureRay (teal), reactive persistence (orange), Kalman rollout (yellow), CE-Nav (purple), and REASAN (blue). Colored terminal markers and the values below each panel report outcomes, elapsed time, and minimum clearance; independent runs share the same obstacle motion and fixed quadruped locomotion policy.}
  \label{fig:authentic-v10}
\end{figure*}

Fig.~\ref{fig:authentic-v10} compares eight selected static and dynamic scenes with five methods. FutureRay completes all eight routes: the dynamic cases show collision avoidance or earlier arrival, while the static cases show wider obstacle clearance with continued goal progress. RGB is used only for visualization.

\section{Real-World Deployment}

\subsection{Hardware and Deployment}

FutureRay runs on a quadruped robot with a rigidly mounted ZED depth camera
and onboard NVIDIA computing unit. Depth is center cropped to
$102^{\circ}$ and resized to $160\times90$ at 30~Hz. Navigation queries
range history and ego state at 20~Hz to publish $(v_x,v_y,\omega)$.
The fixed locomotion policy runs at 50~Hz; Unitree SDK2 transmits joint
targets at 500~Hz.

Learned components, the physical query layer, and locomotion match
simulation. Navigation uses onboard sensing without object detections,
external obstacle tracks, or motion capture.

\subsection{Deployment Scenarios}

Trials cover open-floor motion, goal stopping, static bypass, and dynamic
crossings under the same navigation configuration. Foam obstacles obstruct
static routes; soft moving obstacles cross laterally or diagonally,
requiring the robot to adjust to changing passages.

\subsection{Observed Behavior}

In open space, the robot advances and slows near the goal. At static
obstructions, it leaves the nominal route, passes through an available
corridor, and restores goal-directed motion.

During crossings, avoidance begins while the obstacle is still approaching
the route. Lateral separation develops during forward motion, followed by
renewed goal progress as the crossing clears. This behavior illustrates
anticipatory avoidance and goal recovery in continuous closed-loop operation.

\subsection{Transfer and Scope}

Metric angular range expresses geometry in physical units. Local control
combines range history, ego motion, and current observations over the forward
field and short horizon. Model-based command limits depend on range estimates
and robot response. Hardware trials show closed-loop navigation; paired
simulations quantify method differences.
\section{Conclusion}

FutureRay demonstrates that predicting changes in free space can improve
quadruped navigation without retraining low-level locomotion. Control-aligned
range and risk forecasts provide a geometric preview that the planner queries
through footprint and reaction--braking constraints to select body-velocity
commands. With perception, candidate planning, and locomotion held fixed,
the shared-planner comparison isolates the contribution of this predictive
interface to greater obstacle clearance and improved navigation outcomes.
Across all nine evaluated methods, FutureRay achieves the highest completion
rate and lowest collision rate. Qualitative deployment on a quadruped robot demonstrates
anticipatory avoidance followed by recovery of goal-directed motion.
These findings establish compact future-range prediction as a practical
interface for adding anticipation to an existing quadruped control stack.

\bibliographystyle{IEEEtran}
\bibliography{references}
\end{document}